\documentclass[10pt, a4paper]{article}

\usepackage{lrec-coling2024}

\name{Mitja Nikolaus$^{1}$\thanks{$^1$Work performed while at Aix-Marseille University.}, Abhishek Agrawal$^{2}$, Petros Kaklamanis$^{3}$, Alex Warstadt$^{4}$, \\ \textbf{\fontsize{12}{12}\selectfont Abdellah Fourtassi$^{2}$}} 

\address{$^1$CerCo, CNRS, Toulouse, France\\
        $^2$Aix Marseille Univ, Université de Toulon, CNRS, LIS, Marseille, France\\
        $^3$University of Pennsylvania \\
        $^4$ETH Zürich \\
        mitja.nikolaus@cnrs.fr}

\usepackage{natbib}
\usepackage{multibib}
\makeatletter
\def\@mb@citenamelist{cite,citep,citet,citealp,citealt,citepalias,citetalias}
\makeatother
\newcites{languageresource}{~}

\usepackage{graphicx}
\usepackage{tabularx}
\usepackage{booktabs}
\usepackage{soul}
\usepackage{array}
\usepackage{amsmath}

\usepackage{threeparttable}
\usepackage{multirow}

\usepackage{hyperref}
\usepackage{xstring}

\usepackage{color}

\renewcommand{\cite}{\citep}

\title{Automatic Annotation of Grammaticality in Child-Caregiver Conversations}

\abstract{
The acquisition of grammar has been a central question to adjudicate between theories of language acquisition. In order to conduct faster, more reproducible, and larger-scale corpus studies on grammaticality in child-caregiver conversations, tools for automatic annotation can offer an effective alternative to tedious manual annotation. We propose a coding scheme for context-dependent grammaticality in child-caregiver conversations and annotate more than 4,000 utterances from a large corpus of transcribed conversations. Based on these annotations, we train and evaluate a range of NLP models. Our results show that fine-tuned Transformer-based models perform best, achieving human inter-annotation agreement levels.
As a first application and sanity check of this tool, we use the trained models to annotate a corpus almost two orders of magnitude larger than the manually annotated data and verify that children's grammaticality shows a steady increase with age.
This work contributes to the growing literature on applying state-of-the-art NLP methods to help study child language acquisition at scale.
 \\ \newline  }

\begin{document}

\maketitleabstract

\section{Introduction}

The acquisition of grammar has historically been a central point regarding discussions on the learnability of language from limited input \cite{chomsky_syntactic_1957, gold_language_1967,harris_linguistics_1993,brown_first_1973,piantadosi_modern_2023}.  Traditionally, observational studies on the acquisition of grammar have relied on manual annotations of early child talk. In some cases, notably the question of presence and effectiveness of caregiver corrections following a child's grammatical mistake, research has led to mixed (if not conflicting) results\cite{brown_derivational_1970,nelson_syntax_1973,demetras_feedback_1986,marcus_negative_1993,morgan_negative_1995,saxton_negative_2000,chouinard_adult_2003}.
The lack of consensus can be attributed, at least partly, to the limited sample size used in each study. 

In the current work, we introduce automatic coding as a way forward to address this issue and to help researchers achieve more conclusive results. First, we develop a general coding scheme for the annotation of grammaticality in child-caregiver conversations. Then, we annotate a sample of such conversations to train and evaluate models for automatic annotation, which we use to annotate a large-scale corpus, almost two orders of magnitude larger than the size of the data we coded manually. The developed tools can help researchers perform more cumulative and larger-scale analyses on the development of grammaticality in early childhood and even help adjudicate between general theories of language acquisition \cite{tomasello_constructing_2003,clark_first_2016}.

Our approach differs from typical work on modeling grammaticality using NLP tools, including for research that deals with the linguistic production of adult speakers. While a large portion of this research has dealt with grammaticality (or \textit{acceptability}) of sentences in isolation \cite{lau_grammaticality_2017,warstadt_blimp_2020,warstadt_neural_2019,huebner_babyberta_2021}, here we study the \textit{grammaticality of utterances in conversations}. This covers a differently distributed set of grammatical phenomena (e.g. high proportion of omission errors), and, more importantly, the utterances are often elliptical, i.e., their interpretation depends on the conversational context.

\paragraph{Contributions of this work}
This work makes several contributions.  First, we propose a new coding scheme for the annotation of grammaticality in child-caregiver conversations, based on which we annotate more than 4,000 utterances from English CHILDES \citeplanguageresource{macwhinney_childes_2000}. Additionally, we annotate the specific error category for each ungrammatical utterance.  

Based on this data, we train state-of-the-art NLP models to automatically annotate the grammaticality of utterances and find that the performance of the best models is almost on par with human inter-annotation agreement scores. 

Finally, we use the trained models to annotate all transcripts from English-language CHILDES of children aged 2 to 5 years, which allows us to characterize the developmental trajectory based on this large and diverse corpus.

Our models and annotations, as well as the code for all experiments described in the paper are publicly available at \url{https://github.com/mitjanikolaus/childes-grammaticality}.

\section{Related Work}

\subsection{Automatic Annotation of Grammaticality}\label{sec:related_work_grammaticality}

Supervised approaches for the automatic annotation of grammaticality have often relied on data produced by linguists, e.g., example sentences scraped from linguistics publications \cite{warstadt_neural_2019,trotta_monolingual_2021,mikhailov_rucola_2022,someya_jcola_2023}, often including textbooks \cite[e.g.][]{adger_core_2003,kim_english_2008,sportiche_introduction_2013}.
Using such datasets, early modeling approaches relied on techniques such as n-grams and recurrent neural networks \cite{wagner_judging_2009,lawrence_natural_2000}. Notably, \citet{lau_grammaticality_2017} additionally controlled for confounding factors such as sentence length and lexical frequency to obtain a better classification performance. More recently, the use of large language models pre-trained on large text corpora has enabled substantial performance improvements as measured by comprehensive evaluation benchmarks \cite{warstadt_neural_2019,warstadt_blimp_2020}, with the best Transformer-based models achieving scores that are comparable to human inter-annotation agreement \cite[e.g.][]{he_debertav3_2022}.

Here, we examine whether this progress in the study of isolated sentences can be extended to children's talk in a conversational context, requiring the models not only to adapt to children's data but also to take into account the conversational \textit{context} to evaluate the grammaticality of a given utterance.

\subsection{Automatic Annotation of Children's Grammaticality in Conversation}

Research on automatic annotation of children's productive language in naturalistic conversation has not always focused on grammaticality per se, but instead on other -- more readily automatized measures --  such as Mean Length of Utterances \cite[MLU;][]{brown_first_1973}. 

For the specific measurement of grammatical development, \citet{scarborough_index_1990} proposed the Index of Productive Syntax (IPSyn), in which children are evaluated on how many different syntactic and morphological structures they are correctly producing. Calculating the IPSyn requires the manual scanning of a sample of 100 transcribed utterances for the presence of 56 syntactic and morphological forms.
%evaluating the grammatical complexity
\citet{sagae_automatic_2005} proposed a method to speed up the calculation of IPSyn scores using tools for automatic annotation: The output of a statistical dependency parser was used to narrow down the set of sentences where certain structures may be found by manual annotators.
While such \mbox{(semi-)}automatic methods can provide us with a general estimate of the linguistic productivity of a child, they do not allow for detailed analyses of grammatical phenomena in a conversational context, or per-utterance analyses.

More recently, \citet{hiller_data-driven_2016} focused on the specific case of subject omissions using automated annotation. Based on a small set of hand-annotated data, they trained a Support Vector Machine (SVM) to detect subject omissions. They applied it to perform analyses on a substantially larger set of data. In contrast to this previous work, here we developed a more general characterization of the grammaticality of children's utterances in conversation, including subject omissions but also a dozen more error categories. Our models can be used to obtain a general measure of the grammaticality of utterances as well as for  calculating the overall grammatical competence of a child. They can also be used as a starting point to investigate various mechanisms of language learning, such as corrective feedback \cite{brown_derivational_1970} and communicative feedback \cite{nikolaus_communicative_2023}.

\section{Manual Annotation}

\subsection{Annotation Scheme}

\subsubsection{Grammaticality of Children's Utterances in Conversation}

We develop an annotation scheme adapted for the study of grammaticality of children's utterances in English-language child-caregiver conversations. Based on transcripts of conversations, each child utterance that consists of at least two words is classified as either \textbf{\texttt{grammatical}}, \textbf{\texttt{ambiguous}}, or \textbf{\texttt{ungrammatical}}.\footnote{We exclude all utterances that are unintelligible or not speech-related, such as babbling and other vocalizations like crying or laughing.}
Utterances are annotated as \textbf{\texttt{ungrammatical}} if they contain at least one grammatical error.

\begin{table*}[hbtp]
    \centering \small
    \begin{threeparttable}[b]
\rowcolors{2}{gray!25}{white}
\begin{tabular}{p{.04\textwidth}>{\hangindent=2em}p{.47\textwidth}>{\hangindent=2em}p{.41\textwidth}}
\toprule
\textbf{Label} &\textbf{Cases} & \textbf{Examples} \\
\midrule
 & Ellipses with missing verb or determiner &
``Cookie Monster.'', ``Lunch.'', ``No shoes.''
\\ 
\cellcolor[gray]{1}& Ellipses with missing object &
''I want.'', ``He gave.''
\\ 
\cellcolor[gray]{1}& Ellipses with missing subject, if the context (or verb) clearly points to non-imperative use &
``Want to go to the cinema!''
\\ 
\cellcolor[gray]{1}&SVO/SV questions (except if they are used as clarification requests or to express surprise)
&``You are coming (to the house)?''
\\ 
\multirow[t]{-5}{=}[2.7em]{\cellcolor[gray]{1}
\rotatebox[origin=r]{90}{\texttt{\textbf{ungrammatical}}}
}&Ellipses due to the child being interrupted\tnote{a}
&
``I gave.'' - ``That’s great!''
\\
\midrule

\cellcolor[gray]{1}& Onomatopoeia &
``Miaow miaow'', ``Muuh muh!'', ``Vroom vroom''
\\ 
& Unknown words or vocalizations, baby language, family-internal expression, words spontaneously invented by the child 
&
``Let’s go to the cagriotafer!'', ``eh eh.'' 
\\ 
\cellcolor[gray]{1}&Noun phrases that might be grammatical if accompanied with an appropriate gesture, e.g. a pointing gesture towards an object
&
''A zebra!''\tnote{b}
, ``For the zebra''
\\ 
&Ellipses with missing subject, for which the context does not clearly discriminate between imperative and declarative use&
``Do this.''
\\ 
\cellcolor[gray]{1}&Utterances that are grammatically correct, but not aligned with what the child actually intended to say &
``That’s a nice cup.'' - ``Is it you?''\tnote{c},\newline
``Hide the table!''\tnote{d}
\\ 
&Reciting, Singing, Counting &
``Sweep, sweep, sweep!'', ``One, two, three.''
\\ 
\cellcolor[gray]{1}&Utterances that are strictly ungrammatical but very commonly used in spoken Standard American or British English
&
``Don’t know'', ''You like this?'', ''That all?''
% ''Know what he did?''
% ''Remember what we did last Friday''?
% ``Smells nice.''
% ``Sounds good.''
% ``Where’s your boxes?''
% ``How many women are there?'' - ``There’s two.''
\\ 
\multirow[t]{-8}{=}[5em]{\rotatebox[origin=r]{90}{\texttt{\textbf{ambiguous}}}
}
&Transcription errors &
He like’s animals.
\\

\midrule

\cellcolor[gray]{1}

& Utterances with missing subject, if they are clearly used as imperatives& Look for it!'', ``Take this.'' \\ 
&Utterances with self-repetitions, disfluencies &
``I like I like this.'',
``This is uhm a table.''
\\ 
\cellcolor[gray]{1}&Self-corrections/Reformulations (if the final reformulated utterance is grammatical)\tnote{e} &
``He want she wants a flower!'', \newline ``She is she was very happy''
\\ 
&Exclamations, backchannels &
``Uh oh.'', ``Mhm hm.'', ``All right.'', ``Oh no!''
\\ 
\cellcolor[gray]{1}&Self-repetitions over multiple utterances (both utterances should be marked as grammatical)&
``I want an apple.'' - ``An apple.'' 
\\ 
&Repetitions from the previous utterance (except if the child is repeating an erroneous part from a previous utterance)\tnote{f}
&
 ``It is very hot.'' - ``Very hot.''\newline
 ``This is my funny hat'' - ``My funny hat.''
\\ 
\cellcolor[gray]{1}&Ellipses that are valid responses to a question\tnote{g}&
``Who is that?'' - ``Cookie Monster!'',\newline
``What's this?'' - ``The pasta that dad made!'', \newline
``Are you an artist?'' - ``I am.''
%``[CAR] Which one is it?'' - ``[CHI] This one.''
% ''[CAR] what are you going to do?'' - ''[CHI] Run up the stairs.''
\\ 
&Greetings, calls for attention&
''Good Morning.'', 
''Mummy, mum!''
\\ 
\cellcolor[gray]{1}&Wrong answers, utterances that are logically/semantically wrong or questionable&
 ``Can you say ‘a rat’?'' - ``A cat!''\newline
 ``The sky is green.''
\\ 
&Completions of previous utterances&
``And then he went'' - ``To the cinema''
\\ 
\cellcolor[gray]{1}&SVO/SV questions that function as clarification requests or express surprise & ``This is big.'' - ``This is big?'' \\ 
&Short forms/ contractions commonly used in spoken English&
``Cause I went to school'', ``You’re sposta go there.'', ``That’s a lotta dogs!'', ``Gotcha!''
%``Watcha gonna do?'', ``I like ‘em'', ``I was runnin’'', 
\\ 
\cellcolor[gray]{1}&Utterances with phonological errors (either because of dialect or pronunciation difficulties of the child)
&
``Sesame Stweet'', ``Dis is a dog'',
``Let’s go srough this once again.''
\\ 
\multirow[t]{-14}{=}[0em]{\rotatebox[origin=r]{90}{\texttt{\textbf{grammatical}}}
}&Ellipses that are clearly accompanied by pointing&
``Oh this!'', ``This one.'', ``These cats.''\\\bottomrule
\end{tabular}

    \caption{Annotation guidelines with example cases for each label.}
    \label{tab:annotation_guidelines}

    \begin{tablenotes}
      \scriptsize
      \item[a] As we do not have access to the timing of the utterances, we do not know whether the child was actually interrupted or they just stopped the utterance before completing it. To be consistent, we mark these cases as ungrammatical.
      \item[b] If the determiner is missing (``zebra!''), the utterance should be marked as \texttt{ungrammatical}.
      \item[c] In this case, the child’s response is grammatical but they most likely intended to say something like ``Is it yours?''.
      \item[d] In this case, the child most probably meant to say ``Hide \textit{under} the table''. ``Hide the table'' is strictly speaking grammatical, but we know that there’s actually a grammatical error (missing preposition) if we can infer from the context what the child actually intended to say.
      \item[e] In the case of reformulations across multiple utterances: ``He want'' - ``She wants a flower'' the first utterance should be marked as ungrammatical, the second one as grammatical.
      \item[f] Repetitions that are e.g. missing a determiner should be marked as ungrammatical ``I like the book'' - ``book.''
      \item[g] Usually, questions that ask for a noun (phrase) still require the response to have an appropriate determiner (``What is this?'' - ``A cat.''). If they are missing the determiner, they should be marked as ungrammatical. However, in case the question directly asks for a concept, a response without a determiner is permitted: ``How do we call this?'' - ``Cat!''.
%\\  &Elliptic follow-up questions:& ``What’s this?'' - ``And that?''
    \end{tablenotes}

    \end{threeparttable}
\end{table*}

The grammaticality of each utterance is judged not only based on the utterance itself, but also on the broader context of the conversation. Many utterances in child-caregiver conversations are non-sentential utterances, with highly context-dependent meanings \cite{fernandez_classifying_2007}. 
Consider the following dialog:

\begin{quote}
    Caregiver: \textit{Here take your coat off.} \newline
    Caregiver: \textit{Where do you wanna put your coat?} \newline
    Child: \textit{On the table.} \newline
    \begin{flushright}
    \vspace{-2em}
    \small{--- MacWhinney corpus, 030526a.cha}
    \end{flushright}
\end{quote}

In this example, when judging the grammaticality of the child utterance in isolation, one could be annotating it as \texttt{ungrammatical} as it is missing subject and verb. However, within the context of the preceding utterance, it should instead be marked as \texttt{grammatical}, as it is a valid response to the preceding question.

In contrast, in the following example, the child's utterance is indeed \texttt{ungrammatical} (missing subject and verb), even when taking into account the conversational context: 
\begin{quote}
    Caregiver: \textit{You can play with them on the table.} \newline
    Child: \textit{Lots lots in here.} \newline
    \begin{flushright}
    \vspace{-2em}
    \small{--- Thomas corpus, 020924.cha}
    \end{flushright}
\end{quote}

For each utterance, the annotators are instructed to take into account the preceding context of the conversation for judging its grammaticality.\footnote{However, annotators are instructed to \textit{not} take into account the \textit{following} context of the conversation after the end of the current child's turn. This decision was made in order not to influence the grammaticality judgments based, for example, on the presence of clarification requests from the caregivers, which could bias the annotator into considering that the child's utterance is grammatical in retrospect.}  
%(Indeed, if the caregiver asks for clarification, it is probably because there is some mistake in the child's utterance).}

The label \texttt{ambiguous} is introduced to cover cases in which the grammaticality depends on context that is impossible to infer from the transcript alone (e.g. information about the visual context) as well as cases in which the concept of grammaticality is not applicable.\footnote{\citet{castilla-earls_impact_2020} introduced a class of \texttt{ambiguous} utterances in addition to \texttt{grammatical} and \texttt{ungrammatical} for similar reasons.} For example, the utterance \textit{``do this.''} could be grammatical as an imperative. It could also be a case of a subject omission error if the child actually intended to say \textit{``I do this''}. In some cases, but not always, it is possible to infer the intended meaning from the context of the conversation (e.g. if the preceding utterance is \textit{``Who does this?''}, it is most likely a case of subject omission). 

Utterances that only consist of a noun phrase are annotated as \texttt{ungrammatical} (as they are missing a finite verb), except if they function as responses to questions (``What is this?'' - ``An apple.''). Another exception is the case of an isolated noun phrase that can function as a request for attention if accompanied by an appropriate gesture, such as pointing towards an object (e.g. \textit{``A zebra!''}). As we do not have access to the visual context from the transcribed conversations, such utterances are annotated as \texttt{ambiguous}.
More example cases for each label can be found in Table~\ref{tab:annotation_guidelines}.

\subsubsection{Grammatical Error Categories}

\begin{table*}[hbt]
    \centering \small
    \rowcolors{2}{gray!25}{white}

\begin{tabular}{p{.11\textwidth}p{.13\textwidth}>{\hangindent=2em}p{.27\textwidth}p{.26\textwidth}p{0.11\textwidth}}
\toprule
\textbf{Category (broad)} &\textbf{Category \newline
(fine-grained)} & \textbf{Description} & \textbf{Examples} & \textbf{Frequency (Number)} \\
\midrule
\multirow[t]{3}{=}{Syntactic } & 
\texttt{subject} & Missing subject & 
``Is hot?'', 
``Going there.''
& 17.7\% (322)
\\ 
\cellcolor[gray]{1}& \texttt{object} & Missing object 
& ``Can we look for.'',
%``She says.''\newline
``I like.'' 
& 6.4\% (116) 
\\ 
& \texttt{verb} & Missing verb (incl. copula)
&
``This yours.'',
%``Daddy raisins.''\newline
``Because it.''
& 14.7\% (267) 
\\
\midrule
\multirow[t]{2}{=}{Noun morphology }
\cellcolor[gray]{1} &
\texttt{possessive} & Missing or wrong use of possessive 
& ``What's the other boy name?''\newline
``Where is Julia house?''
&  1.4\% (26) 
\\ 
& 
\texttt{plural} & Wrong plural form or use & 
``No I like mans.'',
``More truck.'' 
&  0.9\% (17)
\\
\midrule
\multirow[t]{2}{=}{Verb morphology }
\cellcolor[gray]{1} & \texttt{sv$\_$agreement} & Subject-verb agreement errors %(incl. 3rd person singular s, to be)
& 
``He want cake.'',
``She are happy!'' 
&  3.4\% (61)
\\ 
& 
\texttt{tense\_aspect} & Wrong tense or aspect inflection of a verb %(excluding errors with present progressive, these have a separate label)
&
%``Did I did it?''\newline
``He's forgot me.'', 
``She falled over.''
&  7.9\% (143)
\\
\midrule
\multirow[t]{4}{=}{Unbound morphology} 
\cellcolor[gray]{1} & \texttt{determiner} & Missing or wrong determiner & 
``Blue wheel.'',
%``See window?''\newline
``A ice cream?''
& 18.8\% (342)
\\ 
& \texttt{preposition} & Missing or wrong preposition &
``I want see it.'',
``Give it me!''
& 4.1\% (75)
\\ 
\cellcolor[gray]{1}& \texttt{auxiliary} & Missing or wrong auxiliary verb &
``We not to put them away.'',
``Someone been crashed.'' 
& 11.4\% (207)
\\ 
& \texttt{present$\_$\newline progressive} & Wrong present progressive form& ``It coming.'', ``What's he say?''
& 4.3\% (78)
\\
\midrule
\cellcolor[gray]{1} Other
& \texttt{other} & Any other kind of grammatical error &
``Many money!'' (many/much)\newline
``Why it's falling?'' (word order)\newline
``My want to eat'' (wrong case)
& 8.9\% (162)
\\
\bottomrule
\end{tabular}
    \caption{Descriptions of error categories that are used to label ungrammatical utterances. The last column is indicating the frequencies (and number of occurrences) calculated from our manual annotations.}
    \label{tab:error_categories}
\end{table*}

For analysis purposes, we additionally annotate the fine-grained \textit{types} of errors for each \texttt{ungrammatical} utterance. The coding scheme is slightly adapted from \citet{hiller_data-driven_2016} and \citet{saxton_prompt_2005}.\footnote{We do not distinguish errors of omission/insertion/substitution (all errors in the \texttt{subject}, \texttt{verb}, and \texttt{object} categories are categorized as errors of omission.) We group regular and irregular past tense errors in the group \texttt{tense\_aspect} (thereby also including errors with, e.g. participles). Further, regular and irregular plural errors are merged and all kinds of subject-verb agreement errors %(third-person ``s'', wrong use of is/are) 
are included in the group \texttt{sv\_agreement}).}
Table \ref{tab:error_categories} describes all error categories along with specific examples. 
An utterance can be assigned multiple error categories, if appropriate.

\subsection{Data}\label{sec:data}

Transcribed conversations are taken from English CHILDES \citeplanguageresource{macwhinney_childes_2000} from children between 2 and 5 years of age. Transcripts are randomly selected from the available corpora, in order to increase variability of conversational contexts, parenting styles, and socioeconomic status.\footnote{As our coding scheme was developed for Standard American and British English, we filter the data for diverging dialects. Fine-grained dialect information is not typically available in CHILDES, so we identify cases to be excluded by searching for caregivers whose speech contains a substantial number of indicative bigrams (``she don't'', ``you was'') and exclude the corresponding corpora.}

All transcripts are concatenated and then split to create files that each contain exactly 200 children's utterances. In total, 21 files are annotated, resulting in \textbf{4200 annotated utterances}.

\subsection{Manual Annotation Results}\label{sec:manual_annotation_results}

The annotations are performed by 3 annotators. For the first 12 files, the annotations are discussed after each file in order to reach sufficient agreement on the annotation scheme. Each label for which at least 2 annotators disagreed is discussed until a consensus is established. From file 13 on, agreements are not discussed, and final labels are calculated as the majority vote from the 3 annotators. For these files, the \textbf{inter-annotation agreement is 0.76} (Krippendorff's Alpha, with ordinal level of measurement \cite{krippendorff_content_2018}).\footnote{Cohen's kappa across the three annotators is on average 0.72 (standard deviation: 0.03).}

In total, 1333 (32\%) utterances are annotated as \texttt{ungrammatical}, 648 (15\%) as \texttt{ambiguous}, and 2219 (53\%) as \texttt{grammatical}.
For all \texttt{ungrammatical} utterances, additional fine-grained error categories (cf. Table~\ref{tab:error_categories}) are added by one annotator.
Their distribution is included in the last column of Table \ref{tab:error_categories}.

\section{Automatic Annotation}

\subsection{Models}

Based on our survey of the literature on automatic annotation of grammaticality (Section \ref{sec:related_work_grammaticality}), we select a range of baseline models and state-of-the-art Transformer-based models for comparison.

We train the models to classify utterances as \texttt{grammatical}, \texttt{ungrammatical}, or \texttt{ambiguous} based on the annotations presented in Section \ref{sec:manual_annotation_results}.
We run a majority classifier, SVMs based on n-gram features, and an LSTM \cite{hochreiter_long_1997} that we pre-train on English CHILDES using a language modeling objective and fine-tune on the task. Further, we fine-tune on the grammatically task the following pre-trained Transformer models: BERT (bert-base-uncased) \cite{devlin_bert_2019}, GPT2 \cite{radford_language_2019}, RoBERTa (roberta-large) \cite{liu_roberta_2019}, and DeBERTa (deberta-v3-large) \cite{he_deberta_2020,he_debertav3_2022}. The LSTM as well as the Transformer models are provided with a list of preceding utterances as conversational context in addition to the target utterance (see also Section \ref{sec:effect_context_length}).
We use early stopping by measuring Pearson's Correlation Coefficient (PCC)\footnote{Related work on grammaticality classification usually relies on Matthews' Correlation Coefficient \cite{matthews_comparison_1975}; we use PCC as it takes into account the fact that we have 3 classes, which are ordinal.} on a validation set (20\% of the training data) to avoid over-fitting during the fine-tuning. Further, we counteract the problem of imbalanced classes (cf. Section \ref{sec:manual_annotation_results}) by applying class weights on the loss.
Further implementation details on the models can be found in Appendix \ref{sec:app_model_training_details}.

\subsection{Results}

We evaluate the models using 5-fold cross-validation, while making sure that there are no transcripts overlapping between training and test sets. As evaluation metrics, we report mean and standard deviation (over the 5 cross-validation folds) of Accuracy as well as PCC.

For models taking into account conversational context, we use a context length of 8 preceding turns. We base this decision on experiments with DeBERTa showing that this context length is optimal for that model (cf. Section \ref{sec:effect_context_length}).

To have an estimate of how the models perform in comparison to inter-annotation agreement, we calculate the same evaluation metrics for human annotators. We report the mean and standard deviation of the pairwise Accuracy and PCC scores across the three annotators.

Table \ref{tab:results} shows the results. Regarding the evaluation metrics, we clearly see the advantage of using the PCC score over Accuracy; the latter tends to -- misleadingly -- favor classifiers with a majority-class bias. For example, Accuracy shows only a minimal performance difference of a majority class classifier compared to the SVM classifiers, while their PCC scores differ substantially.
When comparing PCC scores, we observe that the SVMs show increasing performance with increasing $n$ of their n-gram features, but reaching ceiling starting from 5-grams. The LSTM performs slightly worse than the SVMs according to PCC, and slightly better in Accuracy. The fine-tuned large language models outperform these models by a large margin, with DeBERTa performing best. The PCC score of the best models is very close to human annotators' agreement (0.71 vs. 0.76).

\begin{table}[!ht]
\begin{center}

\begin{tabular}{lll}
\toprule
model & PCC & Accuracy \\
\midrule
Majority class & 0.00 $^{\pm0.00}$ & 0.53 $^{\pm0.11}$ \\
SVM (1-gram) & 0.28 $^{\pm0.09}$ & 0.55 $^{\pm0.02}$ \\
SVM (2-gram) & 0.29 $^{\pm0.09}$ & 0.56 $^{\pm0.03}$ \\
SVM (3-gram) & 0.30 $^{\pm0.08}$ & 0.56 $^{\pm0.03}$ \\
SVM (4-gram) & 0.31 $^{\pm0.08}$ & 0.56 $^{\pm0.03}$ \\
SVM (5-gram) & 0.32 $^{\pm0.08}$ & 0.56 $^{\pm0.03}$ \\
SVM (6-gram) & 0.31 $^{\pm0.08}$ & 0.55 $^{\pm0.03}$ \\
LSTM & 0.27 $^{\pm0.17}$ & 0.58 $^{\pm0.07}$ \\
GPT2 & 0.50 $^{\pm0.10}$ & 0.69 $^{\pm0.04}$ \\
BERT & 0.63 $^{\pm0.07}$ & 0.73 $^{\pm0.04}$ \\
RoBERTa & 0.70 $^{\pm0.07}$ & 0.79 $^{\pm0.04}$ \\
DeBERTa & 0.71 $^{\pm0.05}$ & 0.77 $^{\pm0.03}$ \\
\midrule
Human annotators & 0.76 $^{\pm0.04}$ & 0.80 $^{\pm0.02}$ \\
\bottomrule
\end{tabular}

\caption{Accuracy and PCC scores on test set. Standard deviation over 5-fold cross-validation with varying model random initializations.}\label{tab:results}
 \end{center}
\end{table}

\subsection{Analyses}

\subsubsection{Effect of Context Length}\label{sec:effect_context_length}

One major contribution of this work is the annotation of grammaticality \textit{in context}, that is, by taking into account the preceding utterances in the conversation. In order to explore to what degree the models benefit from the context, we train the best-performing model (DeBERTa) with varying numbers of preceding utterances as context.

\begin{figure}[htb]
\begin{center}
\includegraphics[width=\columnwidth]{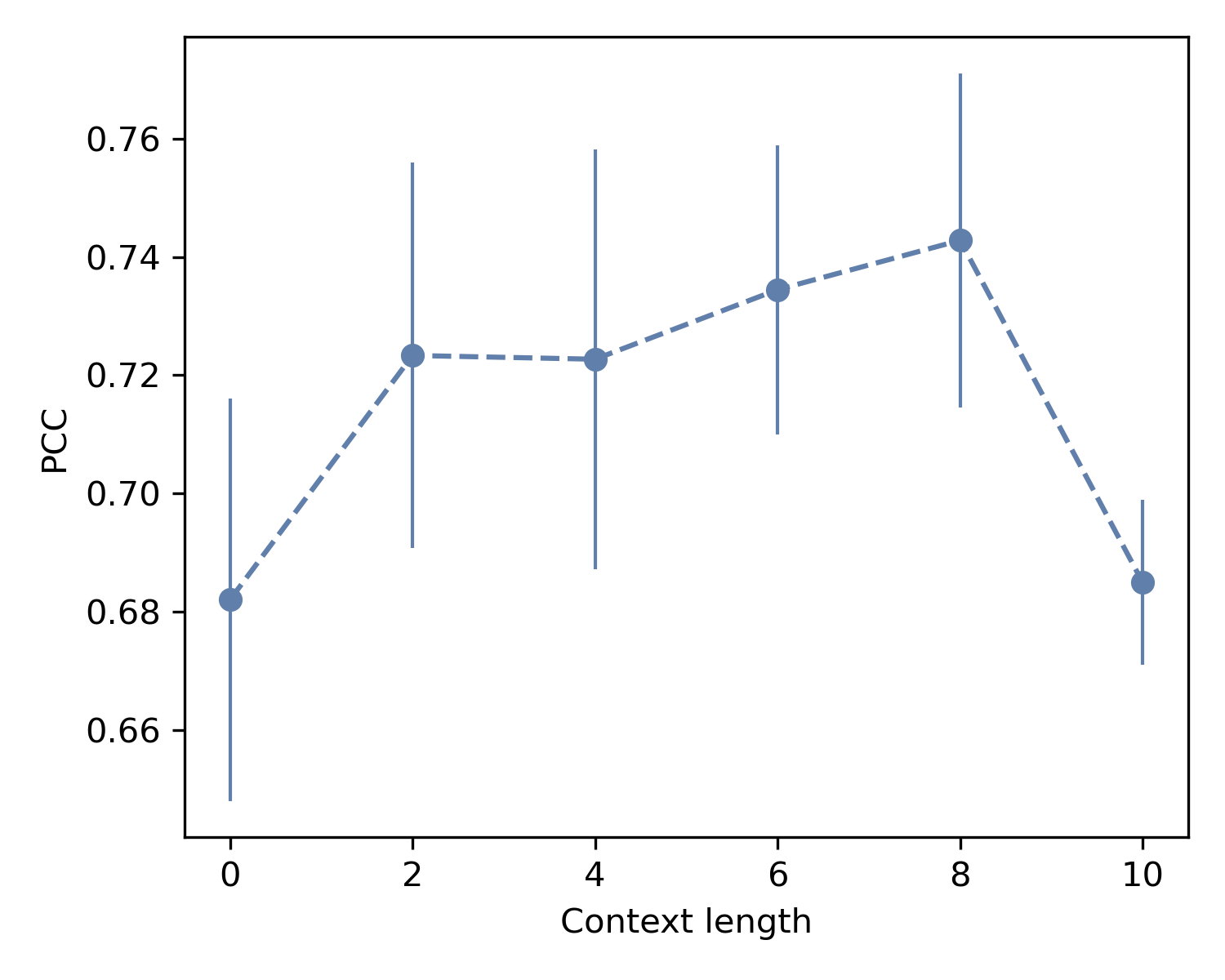} 
\caption{Mean and standard deviation of validation set PCC scores of DeBERTa as a function of the number of preceding utterances in the context.}
\label{fig:context_lengths}
\end{center}
\end{figure}

Figure \ref{fig:context_lengths} shows the PCC scores on the validation set for context lengths 0 to 10. We observe a clear increase in performance for models with 2 utterances in the context as compared to no context (i.e. judging the grammaticality only based on the utterance itself). The performance further increases up to a context length of 8, after which it decreases slightly. We conclude that for this version of DeBERTa, a context length of 8 preceding utterances is optimal.

\subsubsection{Effect of Training Data Size}\label{sec:effect_training_data_size}

Here we explore how the best model (DeBERTa) performs if it is only provided a subset of the training data. Such analyses can provide us insight into the possibilities of further improving model performance by manually annotating additional data. 
We train models using 20\%, 40\%, 60\%, and 80\% of the data. The cross-validation splits and test sets are kept the same.

\begin{figure}[!ht]
\begin{center}
\includegraphics[width=\columnwidth]{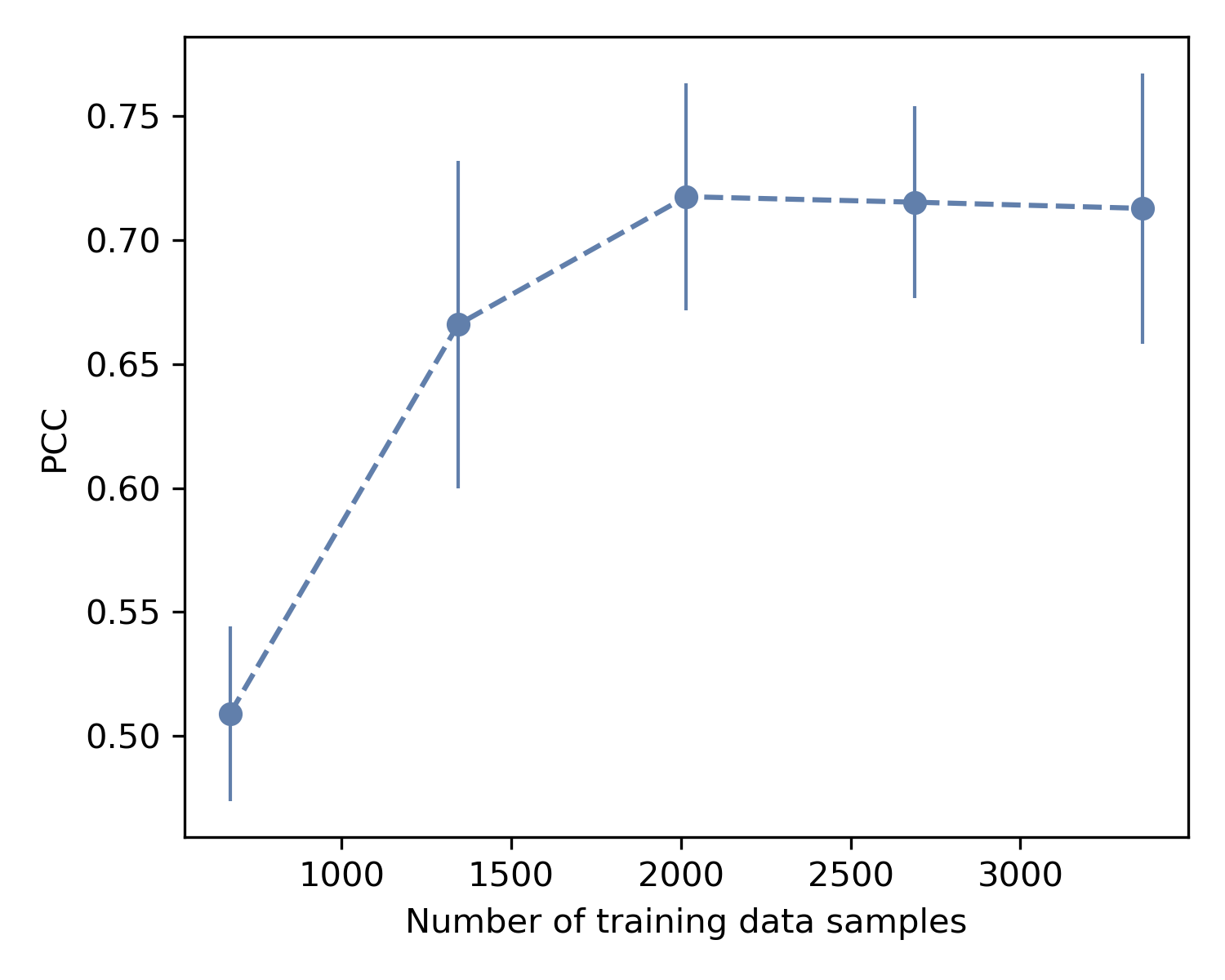} 
\caption{Effect of training data size on test set PCC scores of DeBERTa. The plot displays performance for models trained on 20\%, 40\%, 60\%, 80\%, and 100\% of the training data.}
\label{fig:train_data_size}
\end{center}
\end{figure}

In Figure \ref{fig:train_data_size}, we display model performance as a function of training data size. The curve has a logarithmic-like shape. 
Between a training set size of 1000 and 2000 samples we observe a major improvement in PCC score. Starting from around 2000 training samples the model performance reaches ceiling. We therefore conclude that scaling up our manual annotation efforts is unlikely to lead to substantially improved automatic annotations.

\subsubsection{Error Analysis}\label{sec:error_analysis}

We perform an analysis of errors of the best-performing model (DeBERTa).
Table \ref{tab:confusion_matrix} presents the confusion matrix for the automatic annotations on the test sets (data aggregated from the 5 cross-validation runs).

\begin{table}[!ht]
\begin{center}
\begin{tabular}{lrrr}
\toprule
 & Ungramm. & Ambig. & Gramm. \\
\midrule
Ungramm. & 0.72 & 0.13 & 0.15 \\
Ambig. & 0.17 & 0.56 & 0.27 \\
Gramm. & 0.04 & 0.09 & 0.87 \\
\bottomrule
\end{tabular}
\caption{Confusion matrix for DeBERTa, normalized over the true labels.}\label{tab:confusion_matrix}
 \end{center}
\end{table}

We find that the model commits most errors for the \texttt{ambiguous} class (only 56\% are correctly predicted) and performs best for \texttt{grammatical} utterances (87\% correct). This pattern reflects the number of training examples available for each class (cf. Section \ref{sec:manual_annotation_results}).
Manual inspection of the \texttt{ambiguous} utterances reveals that most misclassified examples are cases of missing subject, for which it is unclear whether they are used as imperative or declarative statements as well as noun phrases with missing verbs, for which the visual context was missing to judge whether there was a pointing gesture towards the mentioned object (see also Table \ref{tab:annotation_guidelines}).

Based on the error category annotations for \texttt{ungrammatical} utterances (Table \ref{tab:error_categories}) we can additionally analyze the model's performance for different kinds of grammatical errors.
\begin{figure}[!ht]
\begin{center}
\includegraphics[width=\columnwidth]{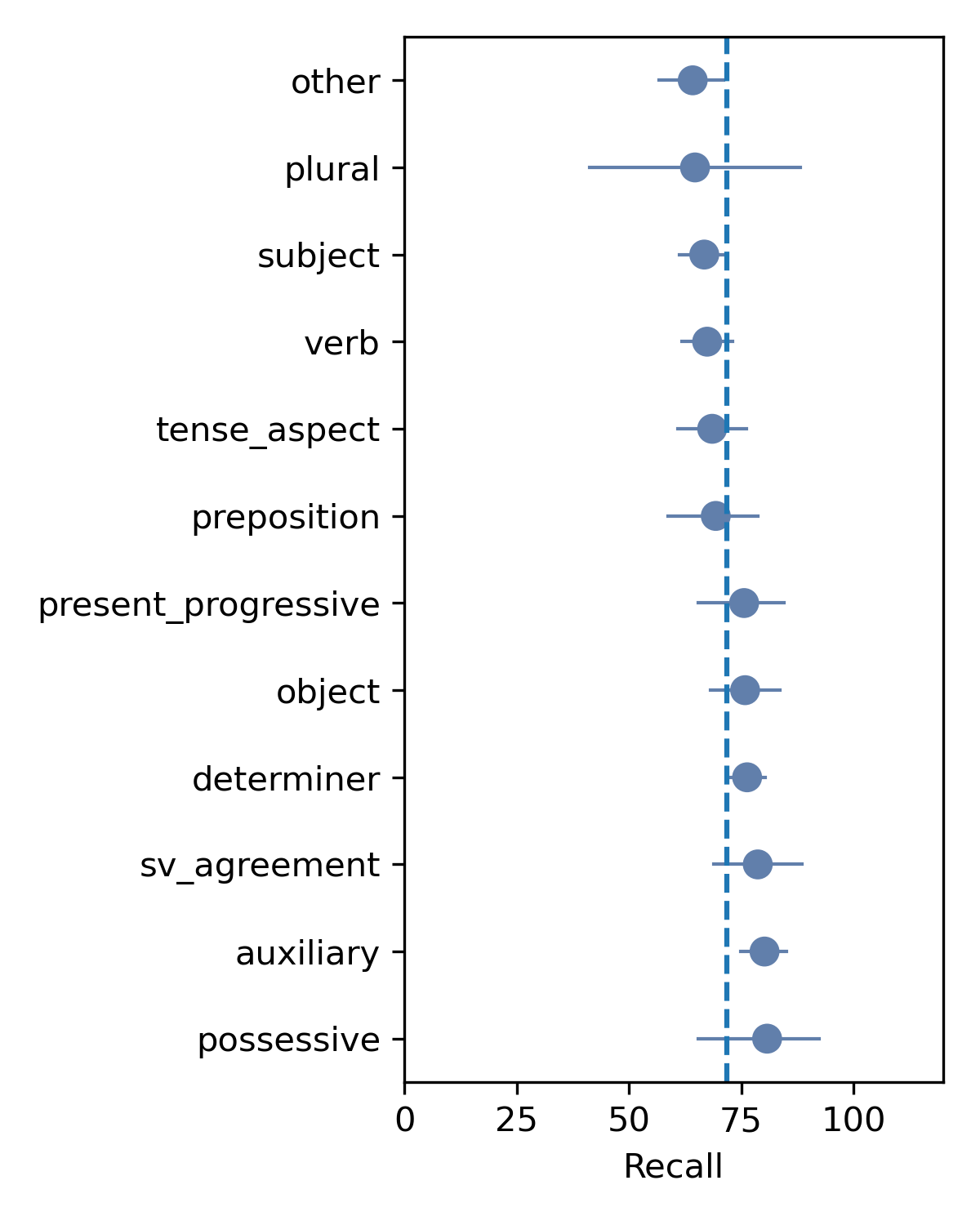} 
\caption{Recall scores for \texttt{ungrammatical} utterances with different error types. Error bars indicate 95\% confidence intervals estimated using bootstrapping. The dotted line indicates the overall average Recall.}
\label{fig:error_analysis}
\end{center}
\end{figure}

Figure \ref{fig:error_analysis} shows the Recall\footnote{We cannot report Precision or F-score as we do not have error category annotations for false positives.} for the \texttt{ungrammatical} class split up by the different error categories. The scores do not diverge much from the average Recall, with the exception of the \texttt{other} and the \texttt{plural} class (lowest Recall). One explanation could be that the \texttt{other} class includes errors from various sources that are rather scarce (errors with case or word order), and therefore hard to learn for the model. The \texttt{plural} class is the least frequent in the training data (only 17 examples).
On the other hand, detecting errors of missing auxiliaries and possessives could be easy as there is a large number of training examples for auxiliaries and the error patterns for both classes are rather consistent (a \texttt{possessive} error usually involves a missing suffixed ``s'').

\begin{figure*}[!htb]
\begin{center}
\includegraphics[width=\textwidth]{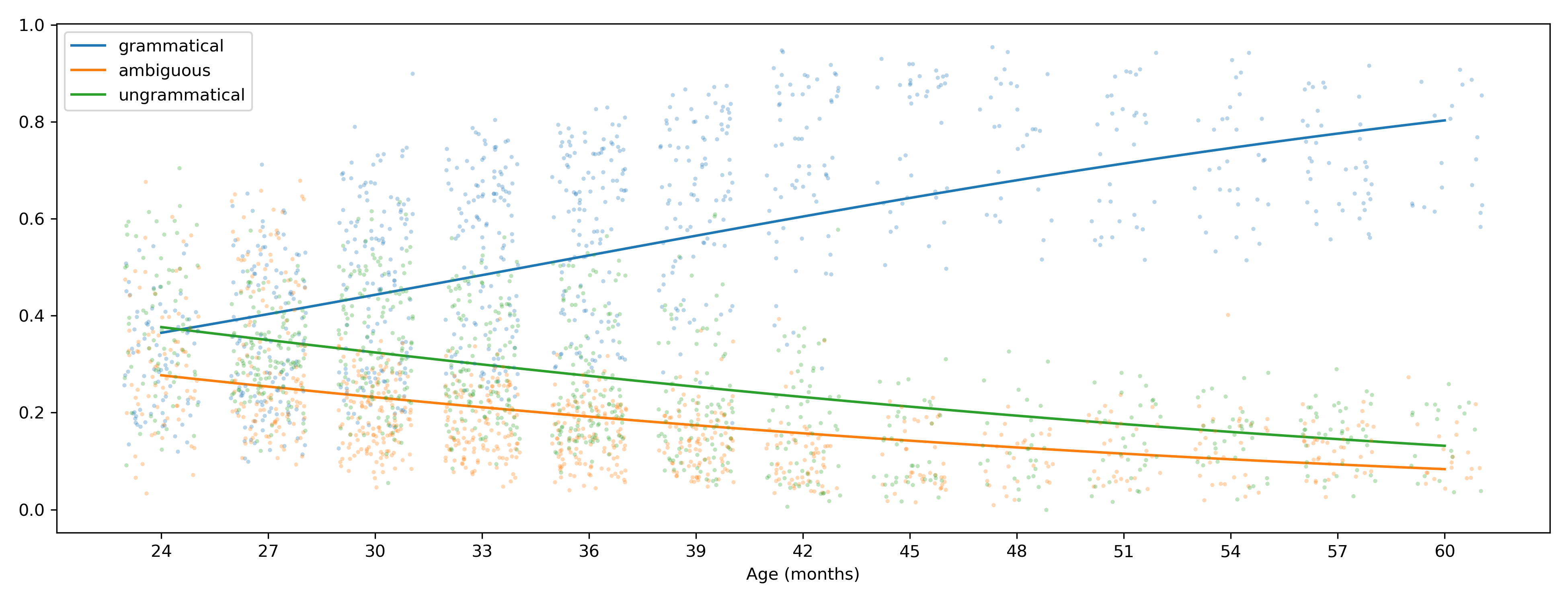} 
\caption{Proportion of \texttt{grammatical}, \texttt{ambiguous}, and \texttt{ungrammatical} utterances for transcripts in English CHILDES of children aged 2 to 5 years. Additionally, we display fitted logistic regression curves.}
\label{fig:annotations_grammaticality_over_age}
\end{center}
\end{figure*}

\section{Large-scale Annotation of English CHILDES}\label{sec:large_scale_annotation}

As a first application and sanity check of the models we introduced in this work, we annotate the grammaticality of all children's utterances in English CHILDES for children aged 2 to 5 years (excluding the manually annotated data). In total, we automatically annotate 276,200 utterances from 321 children and 1900 transcripts. % (that is, an increase by almost two orders of magnitude from the manually annotated data).
To obtain the labels, we calculate the majority vote of all 5 fine-tuned DeBERTa models (there are 5 models trained on the different cross-validation splits).

In Figure \ref{fig:annotations_grammaticality_over_age} we present the proportion of \texttt{grammatical}, \texttt{ambiguous}, and \texttt{ungrammatical} utterances for each annotated transcript.\footnote{We excluded transcripts with less than 100 child utterances to reduce clutter.} Further, we display fitted curves of a logistic regression for each target label.
We observe a clear increase in the proportion of \texttt{grammatical} utterances with increasing age. At the same time, the proportion of \texttt{ambiguous} and \texttt{ungrammatical} utterances decreases.
We use mixed effects models to verify these trends. Regarding the proportion of \texttt{grammatical} utterances we fit the following model:
\begin{equation}
\texttt{grammatical} \sim \texttt{age} +  (\texttt{1} | \texttt{transcript})
\end{equation}
We obtain $\text{age}: \beta = 0.014, SE=0.001, p < 0.001$, indicating a significant positive correlation with age. We run equivalent models for the proportion of \texttt{ambiguous} and \texttt{ungrammatical} utterances and obtain significant negative correlations. For \texttt{ambiguous} utterances: $\text{age}: \beta = -0.006, SE<0.001, p < 0.001$ and \texttt{ungrammatical} utterances: $\text{age}: \beta = -0.008, SE<0.001, p < 0.001$.

\section{Limitations}
In order to close the remaining small performance gap between the models and human annotators, one possibility would be to increase the amount of manual annotations. However, our experiments with varying training data sizes show that model performance most probably won't increase substantially with a simple increase in training data size (Section \ref{sec:effect_training_data_size}). On the other hand, our error analysis reveals that many failure cases are likely caused by imbalanced classes in the training data (Section \ref{sec:error_analysis}). In order to address these issues, future annotation efforts could be targeted to obtain more training data for \texttt{ungrammatical} and \texttt{ambiguous} utterances.

The current annotations allow for a broad classification of utterances into \texttt{grammatical}, \texttt{ungrammatical}, and \texttt{ambiguous}. While this is a reasonable first step for the study of grammaticality, many patterns are dependent on specific error types. For example, the effects of utterance length on grammaticality differ for errors of omission vs.~commission \cite{castilla-earls_complex_2022}. Further, \citet{saxton_negative_2005} found that corrective feedback for syntactic errors is more frequent than for morphological errors, and that negative feedback in the form of reformulations is associated with gains in the grammaticality of child speech for 3 out of 13 tested grammatical error categories.
More generally, we can gain insight from the study of a specific grammatical phenomenon, such as learning of the English past tense \cite{saxton_contrast_1997,rumelhart_learning_1986,marchman_continuity_1994,mcclelland_rules_2002}.
To enable more fine-grained analyses of specific error classes, models could be trained to classify the error type in addition to the general grammaticality. As the distribution of error types is highly skewed (cf. Table \ref{tab:error_categories}), there is currently not enough manually annotated data to train models for a reliable classification. Again, targeted annotations could be carried out to increase the number of examples of less frequent error types.

Another important limitation of our contribution is that our annotations assume children and caregivers speak Standard American or British English. In some cases, sentences that are labeled ungrammatical (e.g., ``I been here.'', ``You was there.'', ``She don't like it.'') are grammatical in other English dialects, and so our annotated data and classifiers are not appropriate for the study of other dialects. Even though we make efforts to filter out corpora of diverging dialects (cf. Section \ref{sec:data}), some instances in the dataset (manually or automatically labeled) may have been missed and, therefore, contain inaccurate labels.

\section{Discussion and Conclusion}

Research in child language acquisition has recently started to move towards large-scale studies and cross-lab collaborations to overcome issues such as small sample sizes, lack of population diversity, and inconsistent measures \cite{frank_collaborative_2017,byers-heinlein_building_2020}. The current work contributes to this ongoing effort in the community, providing a tool for the automatic annotation of grammaticality in child-caregiver conversations. This tool will enable researchers to conduct reproducible and cumulative research on a large scale.

We develop a coding scheme for the annotation of the grammaticality of children's utterances in conversation and manually annotate a representative sample. Based on these annotations, we train and evaluate a range of NLP models on this task. We find that the best models are performing on par with human annotators. 

Much research in NLP has dealt with the annotation of grammaticality of utterances in isolation \cite{warstadt_neural_2019,warstadt_blimp_2020}. Here we deal with grammaticality in naturalistic child-caregiver conversations and highlight important differences. Indeed, one of the main contributions of our work is the finding that the grammaticality of an utterance is dependent on the conversational context.
Analyzing the dependence of model performance on context length (i.e., how many previous utterances are given to a model in order to best judge the grammaticality of a target utterance) revealed that while it is possible to reach decent performance when annotating the grammaticality of utterances in isolation (without context), the addition of two previous utterances from the conversational context results in a substantial improvement. The best performance is reached with a context length of 8 utterances. 

Finally, we show that the developed tool can be used to study the trajectory of grammatical development by applying it to annotate a large-scale corpus, enabling more systematic research into the underlying learning mechanisms. 
%For example, previously used automatized measures for syntactic development such as MLU increase with age in typical development \cite{miller_relation_1981} but have the limitation that they reach ceiling when the children reach 3 years \cite{klee_relation_1985}. We observe an increase in grammaticality with our automatic annotations that continues until at least 5 years of age.

A promising area of application of the proposed models is the study of grammaticality in language impairment. It has been found that children's productive performance in terms of grammaticality is correlated with specific language impairment, and could probably be used as an early indicator of risk \cite{rice_mean_2010,souto_identifying_2014,guo_differentiating_2016,eisenberg_differentiating_2013}.

Additionally, by allowing for more reproducible large-scale investigations, the models can aid in adjudicating debates about the learning mechanisms, such as the debate about the role of the caregiver's corrective feedback in language acquisition \cite{brown_derivational_1970,demetras_feedback_1986,saxton_negative_2000,marcus_negative_1993,morgan_negative_1995,nelson_syntax_1973} as well as providing a more thorough test to newly proposed mechanisms such as communicative feedback \cite{warlaumont_social_2014,nikolaus_communicative_2023}.
%which are inspired by usage-based theories of language acquisition \cite{tomasello_constructing_2003,clark_first_2016}.

\section{Acknowledgements}

This work, carried out within the Labex BLRI (ANR-11LABX-0036) and the Institut Convergence ILCB (ANR-16CONV-0002), has benefited from support from the French government, managed by the French National Agency for Research (ANR) and the Excellence Initiative of Aix-Marseille University (A*MIDEX).

The project leading to this publication has received funding from Excellence Initiative of Aix-Marseille - A*MIDEX (Archimedes Institute AMX-19-IET-009), a French ``Investissements d’Avenir'' Programme.

Further, this work was supported by the ANR MACOMIC (ANR-21-CE28-0005-01).

This work was performed using HPC resources from GENCI–IDRIS (Grant 2022-D011013886).

\clearpage
\section{Bibliographical References}\label{sec:reference}

\bibliographystyle{lrec-coling2024-natbib}
\bibliography{main}

\section{Language Resource References}\label{lr:ref}
\bibliographystylelanguageresource{lrec-coling2024-natbib}
\bibliographylanguageresource{languageresource}

\clearpage

\appendix
\section{Appendix}

\subsection{Model training details} \label{sec:app_model_training_details}

\subsubsection{SVM Classifiers}

We train Support Vector Machines (SVM) based on n-gram features as simple baseline models for the task.

The data is tokenized using byte-pair encoding (BPE) with a vocabulary size of 10,000. Afterwards, for each n-gram level (1-gram, 2-gram, ...), a vocabulary of the 1000 most commonly occurring n-grams in the training set is constructed. The features for a given utterance consists of a sparse array containing the number of occurrences of each n-gram from the vocabulary. For SVMs with n-gram features of $n$ greater than 1, the features from all smaller $n$ are included (for example, the 2-gram model features are 2000-dimensional, consisting of a concatenation of the 2-gram and the unigram features).

These features are fed into a C-Support Vector Classification model with balanced class weights and the default arguments from the implementation in scikit-learn \cite{pedregosa_scikit-learn_2011} which is based on libsvm \cite{chang_libsvm_2011}.

These baseline models are trained without any conversational context.

\subsubsection{LSTM}

This model consists of a single-layer LSTM with 512 hidden units and a maximum sequence length of 200 tokens. 

\paragraph{Pre-training}
As a first step, the LSTM is pre-trained with a language modeling objective on the English CHILDES data (cf. Section \ref{sec:large_scale_annotation}), excluding all the data that is manually annotated (in order not to train on data that will be part of any of the test sets during cross-validation). 
10,000 sentences are set aside as a validation set to perform early-stopping based on the validation loss.

The data is tokenized using (BPE) with a vocabulary size of 10,000 and special speaker tokens for child (\texttt{[CHI]}) and caregiver (\texttt{[CAR]}) that are prepended to each utterance.

The maximum sequence length is set to 200 tokens, the model is trained with a batch size of 100 and Adam optimizer \cite{kingma_adam_2015} with an initial learning rate of $10^{-4}$.

\paragraph{Fine-tuning}
After pre-training, the model is equipped with an additional linear classification layer that is fed the output from the last timestep from the LSTM. Then, it is fine-tuned on the grammaticality classification task using a cross-entropy loss with balanced class weights.

The fine-tuning is also performed using an Adam optimizer with initial learning rate of $10^{-4}$, batch size of 100, and using early stopping based on the PCC score on a held-out validation set (20\% of the training data).

\subsubsection{Transformer-based Models}

We fine-tune the following Transformer-based models: BERT (bert-base-uncased) \cite{devlin_bert_2019}, GPT2 \cite{radford_language_2019}, RoBERTa (roberta-large) \cite{liu_roberta_2019}, and DeBERTa (deberta-v3-large) \cite{he_deberta_2020,he_debertav3_2022}.

We leverage pre-trained models from Huggingface \cite{wolf_transformers_2020}. We prepend special speaker tokens for child (\texttt{[CHI]}) and caregiver (\texttt{[CAR]}) to each utterance.

On top of each respective model, a new linear classification layer is fine-tuned on the grammaticality classification task using a cross-entropy loss with balanced class weights.

This fine-tuning is performed using an AdamW optimizer \cite{loshchilov_decoupled_2018} with initial learning rate of $10^{-5}$, batch size of 100, and using early stopping based on the PCC score on a held-out validation set (20\% of the training data).

% - deberta ctx 6: 1918412

\end{document}